\pdfoutput=1

\documentclass[11pt]{article}
\usepackage{CJKutf8}
\usepackage{float}
\usepackage{placeins} 
\usepackage{graphicx}
\usepackage{EACL2023}
\usepackage{xcolor,pifont}
\newcommand{\hlc}[2][yellow]{{\sethlcolor{#1}\hl{#2}}}
\usepackage{latexsym}
\usepackage{times}  
\usepackage{helvet}  
\usepackage{courier}  
\usepackage{algorithm}
\usepackage{algorithmic}
\usepackage{tikz}
\usepackage{multirow}
\usepackage{amsmath,amssymb,mathtools,bm,etoolbox}
\usepackage{wasysym}
\usepackage{tikz}
\usepackage{xcolor}
\usepackage{soul}

\usepackage{booktabs, multirow, array, makecell, caption}
\usepackage{latexsym}
\usepackage{times}  
\usepackage{helvet}  
\usepackage{courier}  
\usepackage{algorithm}
\usepackage{algorithmic}
\usepackage{tikz}
\usepackage{multirow}
\usepackage{amsmath,amssymb,mathtools,bm,etoolbox}

\DeclarePairedDelimiterX{\ExpArg}[1]{[}{]}{#1}

\usepackage{newfloat}
\usepackage{listings}
\usepackage{natbib}  
\usepackage{caption} 
\usepackage[T1]{fontenc}

\usepackage[utf8]{inputenc}

\usepackage{microtype}

\usepackage{inconsolata}

\title{Revamping Multilingual Agreement Bidirectionally via Switched Back-translation for Multilingual Neural Machine Translation}

\author{
    Hongyuan Lu$^\heartsuit$, Haoyang Huang$^\spadesuit$, Dongdong Zhang$^\spadesuit$,\\
    \textbf{Furu Wei$^\spadesuit$, Wai Lam$^\heartsuit$}\\
    $^\heartsuit$The Chinese University of Hong Kong\\
    $^\spadesuit$Microsoft Corporation\\
    \{hylu,wlam\}@se.cuhk.edu.hk\\
    \{haohua,dozhang,fuwei\}@microsoft.com
}

\begin{document}
\maketitle
\begin{abstract}
Despite the fact that multilingual agreement (MA) has shown its importance for multilingual neural machine translation (MNMT), current methodologies in the field have two shortages: (i) require parallel data between multiple language pairs, which is not always realistic and (ii) optimize the agreement in an ambiguous direction, which hampers the translation performance. We present \textbf{B}idirectional \textbf{M}ultilingual \textbf{A}greement via \textbf{S}witched \textbf{B}ack-\textbf{t}ranslation (\textbf{BMA-SBT}), a novel and universal multilingual agreement framework for fine-tuning pre-trained MNMT models, which (i) exempts the need for aforementioned parallel data by using a novel method called switched BT that creates synthetic text written in another source language using the translation target and (ii) optimizes the agreement bidirectionally with the Kullback-Leibler Divergence loss. Experiments indicate that BMA-SBT clearly improves the strong baselines on the task of MNMT with three benchmarks: TED Talks, News, and Europarl. In-depth analyzes indicate that BMA-SBT brings additive improvements to the conventional BT method.\footnote{Code and data will be available upon publication.}

\end{abstract}

\section{Introduction}
Conventional multilingual neural machine translation (MNMT) leverages independent parallel data during the training process. In comparison, the multilingual agreement (MA) explicitly minimizes the output difference between two source inputs written in different languages but with the same meaning. Despite its success in from-scratch training on MT \citep{magree}, current methodologies suffer from at least two disadvantages that limit their scope of usage. Firstly, conventional MA leverages word alignment tools to create code-switching sentence-level data \citep{magree}. This process usually requires authentic parallel data between multiple language pairs. For example, assuming we would like to enhance Chinese to English and German to English, conventional MA assumes the existence of parallel data from Chinese to German, which however sometimes does not exist. Secondly, the direction of agreement-based learning can be bidirectional \citep{agreement3}, while the direction of conventional multilingual agreement is usually ambiguous. However, since languages usually have different linguistic clues and they are helpful to each other, we argue that optimizing the multilingual agreement explicitly in a bidirectional manner can help the languages to learn from each other and hence further enhance cross-lingual learning. 
\begin{figure}[t!]
\begin{center}
\centerline{
\includegraphics[width=7.5cm]{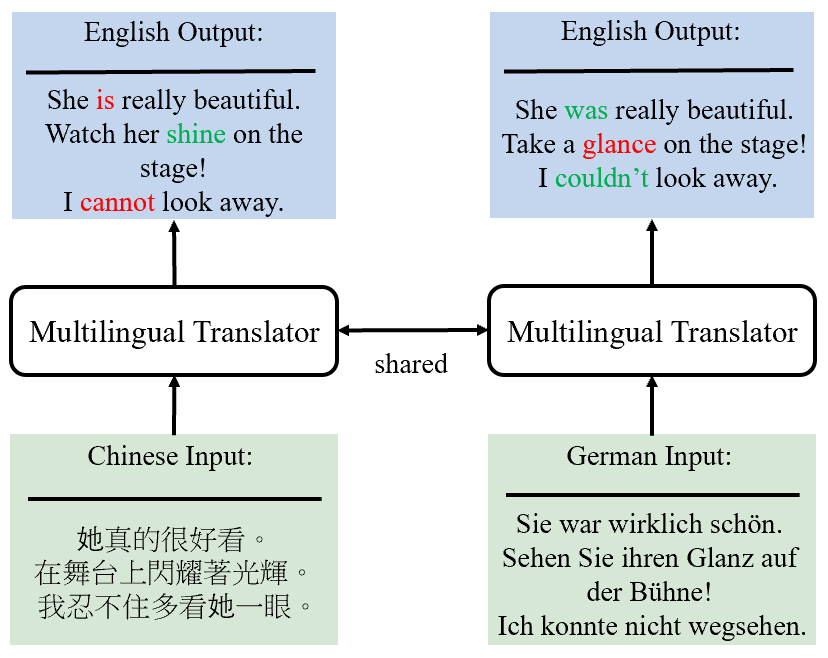}}
\caption{An illustrated example that can be benefited from Multilingual Agreement optimized in a bi-direction manner. The words in green are the correct translation, and the words in red are the wrong translation. Here, Chinese is incorrectly translated since it does not have past tense for verbs, and German is incorrectly translated due to the shared subword unit with different meanings between Glanz (German, shine) and Glance (English, take a brief look at). Best viewed in colour.}
\label{f1}
\end{center}
\vspace{-5mm}
\end{figure}
\par
Figure \ref{f1} depicts such a case that can be benefited from bidirectionally enhanced MA. The underlying reason is that both of the source inputs have cross-lingual ambiguities here. Since Chinese does not have past tense verbs, it is intuitive to use some auxiliary languages with past tense. Furthermore, since German shares partial vocabulary subwords with English under MNMT, this introduces cross-lingual ambiguities and using a language that does not share its vocabulary subwords with English, e.g., such as Chinese, could be helpful.
\par 
As a side note, since MA was proposed as a method for from-scratch training for MT, it was unclear whether conventional MA is also effective as a fine-tuning technique for pre-trained models.
\par 
Furthermore, how to appropriately apply back-translation to a multilingual setting is also an under-studied subject despite its importance.
\par
This paper proposes BMA-SBT, a novel MNMT framework that (i) exempts the need for parallel data between multiple language pairs and (ii) optimizes the MA in a bidirectional manner. To exempt the need for parallel data, we propose switched back-translation to produce synthetic text in some different auxiliary source languages with the translation target.\footnote{For example, Chinese as the source, English as the target, and Japanese as the auxiliary source.} To optimize the MA in an explicit bidirectional manner, we use a bidirectional Kullback–Leibler Divergence loss instead of the code-switching for conventional MA. This enforces the original source language and the synthetic auxiliary language to have the same outputs as the target reference translation in a bidirectional manner.
\par
We conduct experiments on three MT benchmarks: TED Talks \citep{cettolo-etal-2015-iwslt}, News benchmark (News-commentary) and Europarl \citep{koehn-2005-europarl}. Experimental results indicate that BMA-SBT clearly improves the strong pre-trained baselines on all three benchmarks. In-depth analyses indicate that BMA-SBT effectively mitigates cross-lingual ambiguities.
\par
In summary, we make three key contributions:
\begin{itemize}
\setlength\itemsep{0em}
    \item This paper proposes a novel framework called BMA-SBT, the first MNMT framework that achieves MA without the requirement of extra parallel data and explicitly optimizes the MA in a bidirectional manner.
    \item BMA-SBT yields clear improvement on SOTA pre-trained MT model on three MT benchmarks: TED Talks, News, and Europarl.
    \item We conduct in-depth analyses of BMA-SBT. Results indicate that BMA-SBT brings additive improvement to conventional BT and bidirectionality is important for MA.
\end{itemize}
Also, this is the first work that demonstrates the usefulness of MA as a fine-tuning technique.

\begin{figure*}[t!]
\begin{center}
\centerline{
\includegraphics[width=15cm]{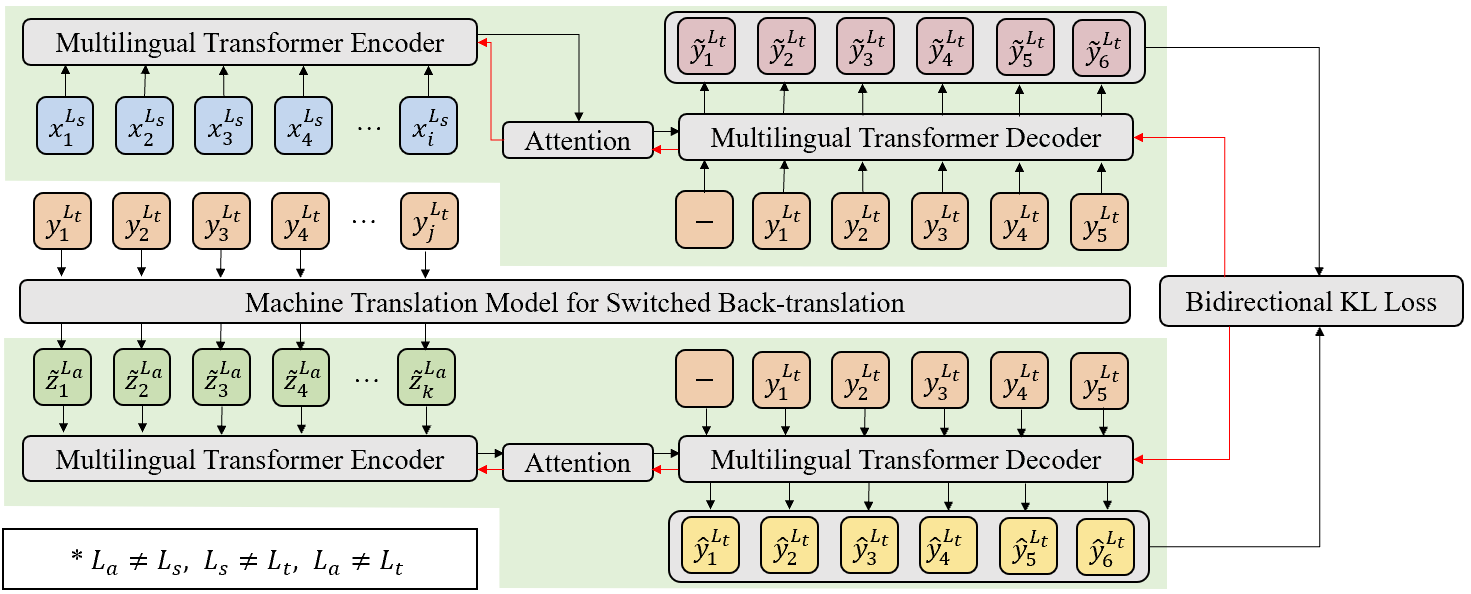}}
\caption{Overview of our proposed BMA-SBT framework. $x$ and $y$ denote the original source and target text written in the source language $L_s$ and target language $L_t$. $\tilde{z}$ denotes the synthetic text translated from the original target text into language $L_a$. $\tilde{y}$ denotes the translation output from the original source text produced by the multilingual Transformer and $\hat{y}$ denotes the translation output from the synthetic text. The letters with subscripts such as $x_i$ denote the $i$-th token in the original source text. The red arrows denote the backward gradient flow computed by the bidirectional KL loss that updates the shared multilingual Transformer encoder and decoder. Best viewed in colour. }
\label{f2}
\end{center}
\end{figure*}

\section{Bidirectional Multilingual Agreement via Switched Back-translation}
\subsection{Multilingual Neural Machine Translation}
We conduct our experiments on the task of MNMT on large-scale pre-trained multilingual translation model \citep{tulg1, 2022arXiv221207752L} that handles multiple languages by sharing a universal subword dictionary among all the languages. For both training and inference, given $I$ languages $\{L_1,,...,L_I\}$, we prefix a special target language token $L_t$ to the source inputs to signal the multilingual model that we are translating from an arbitrary source language to the target language $L_t$.
\par
Given a bilingual dataset for machine translation that consists of $\mathcal{N}$ training instances $\{\mathcal{T}_1,,...,\mathcal{T}_\mathcal{N}\}$, each of the bilingual translation pairs $\mathcal{T}_i$ in the source bilingual dataset $\mathcal{D}_\mathcal{M}$ contains a source input $x$ and the corresponding translation target $y$. With a Seq2Seq generation model \citep{S2S} with parameters $\theta$, we train the model by optimizing the following likelihood: 
\begin{equation}
\label{eq1}
    \mathcal{L}_{main}=\sum_{n=1}^{\mathcal{N}}\mathbb{E}_{x_n,y_n \in \mathcal{D}_{\mathcal{M}}}[-\log P_\theta(y\mid x)],
\end{equation}
where $\mathcal{L}_{main}$ denotes the standard training loss that we adopt for MNMT. 
\subsection{BMA-SBT}
\label{bmabt-label}
In this subsection, we introduce our novel framework \textbf{B}idirectional \textbf{M}ultilingual \textbf{A}greement via \textbf{B}ack-\textbf{t}ranslation (\textbf{BMA-SBT}). Compared to the conventional multilingual agreement, BMA-SBT exempts the need for parallel data and specifies the direction of the multilingual agreement in a bidirectional manner. We first introduce how we use BT to create synthetic parallel data which are appropriate for the use of the multilingual agreement, and we then introduce how to leverage KL divergence loss to make the multilingual agreement bidirectional.
\paragraph{Switched Back-translation} The conventional multilingual agreement (MA) requires authentic parallel data, which could be commonly unrealistic in a real-world setting. Formally, for the translation pair $x$ and $y$ in Equation \ref{eq1}, conventional MA requires another instance $z$, which is written in a different language and in the equivalent meaning to $x$ and $y$. And this process was designed and experimented on from-scratch training. These facts limit the use of the conventional MA. 
\par
To mitigate the above-mentioned shortages, we propose a novel method called switched back-translation that creates synthetic text $\tilde{z}$ written in different source languages by feeding the translation target $y$ to a machine translation model through back-translation.\footnote{While we can use the translation source $x$ to create $\tilde{z}$, we empirically have found that this degrades the improvement. We postulate that if the source text has ambiguities, then this is less helpful to create the auxiliary text with the source text.} Note that $\tilde{z}$, $x$, and $y$ are equivalent in their meanings, but they are written in different languages.
\par 
This helps us to establish a synthetic bilingual auxiliary dataset $\mathcal{D}_{\mathcal{A}}$ that is consisted of $\mathcal{M}$ training instances. We then train the multilingual model by maximising the following likelihood:
\begin{equation}
\label{eq2}
    \mathcal{L}_{auxiliary}=\sum_{n=1}^{\mathcal{M}}\mathbb{E}_{\tilde{z}_n,y_n \in \mathcal{D}_{\mathcal{A}}}[-\log P_\theta(y\mid \tilde{z})].
\end{equation}
\par 
\textit{We also differentiate the switched back-translation we propose here from the conventional BT.} For BT which was originally proposed for bilingual MT \citep{sennrich-etal-2016-improving}, we usually obtain $x'$ from the original monolingual target $y$, where $x'$ should be written in the same source language in our interest. In contrast, BMA-SBT creates $\tilde{z}$ that should have the equivalent meaning as $y$, but it should be written in different languages from both the original source and target languages for the purpose of applying the multilingual agreement.\footnote{For a fair comparison, we use the Baseline Model and the monolingual English sentences in the downstream dataset for data augmentation with BT \citep{sennrich-etal-2016-improving} and SBT.}
\par 
In conclusion, this evolves the conventional MA into a universal fine-tuning technique for MNMT which does not need extra parallel data. BMA-SBT fits the real-world setting and can be applied with some modifications to other generation tasks for cross-lingual learning.
\paragraph{Bidirectional Multilingual Agreement}
The direction for agreement-based learning can be bidirectional \citep{agreement3}. However, the conventional multilingual agreement has an ambiguous direction due to the nature of code-switching. By using parallel data, conventional MA constructs code-switching data $c$ from $x$ and $z$, which denotes the translation source and the authentic auxiliary text respectively. Note that $x$ and $z$ have the same meaning to the translation target $y$, but they are written in different languages. The code-switching is then done with a word alignment tool between $x$ and $z$ at the word level, usually with a low code-switching replacement ratio as low as 10\% \citep{magree}. Formally, conventional MA trains MNMT by maximising the following likelihood: 

\begin{equation}
\label{eq3}
    \mathcal{L}_{MA}=\sum_{n=1}^{\mathcal{Q}}\mathbb{E}_{c,y_n \in \mathcal{D}_{\mathcal{C}}}[-\log P_\theta(y\mid c)],
\end{equation}
where $y$ denotes the translation target, $\mathcal{D}_{\mathcal{C}}$ denotes the code-switching dataset automatically constructed, and $\mathcal{Q}$ denotes the number of samples in the code-switching dataset.\par
In addition to the fact that conventional MA requires authentic data $z$ which is not always realistic, we also argue that code-switching optimizes in an ambiguous direction, usually with a low code-switching ratio as low as 10\%. Therefore, we consider that cross-lingual learning could be less efficient in this manner. As depicted in Figure \ref{f1}, MNMT can be benefited by encouraging multilingual agreement in a bidirectional manner. 
Hence, we use a KL divergence loss to specify the direction of multilingual agreement in a clear bidirectional manner. Since the authentic parallel text $z$ is not always available, we use the aforementioned synthetic auxiliary text $\tilde{z}$ to calculate a bidirectional MA (BMA) divergence loss:
\begin{equation}
\mathcal{L}_{BMA}=\alpha\mathcal{L}_{KL_1}+(1-\alpha)\mathcal{L}_{KL_2},
\end{equation}
\begin{table*}[h!]
\centering
\setlength\tabcolsep{6pt}
\setlength\aboverulesep{0pt}\setlength\belowrulesep{0pt}
\setcellgapes{2.5pt}\makegapedcells
\begin{tabular}{l|cccccc|c}
\hline
\textbf{Model} & \textbf{Fr$\rightarrow$En} & \textbf{De$\rightarrow$En} & \textbf{Zh$\rightarrow$En} & \textbf{Vi$\rightarrow$En} & \textbf{Cs$\rightarrow$En} & \textbf{Th$\rightarrow$En} & \textbf{Avg.}\\
\hline
\multicolumn{8}{l}{\textit{Sentence-level Systems}}
\\
\hline
HAN$\dag$ & - & - & 24.00 &-&-&-&-\\
M2M-100  & 50.18 & 42.24 & 26.62 & 34.92 & 37.84 & 27.28 & 36.51\\
mBART  & 48.69 & 44.80 & 28.39 & 37.18 & 39.47 & - & -  \\
Baseline Model + BT & 50.69 & 47.07 & 30.35 & 39.59 & 43.05 & 32.30 & 40.51\\
\hline
\multicolumn{8}{l}{\textit{Document-level Systems}}\\
\hline
mT5$\dag$ & - & - & 24.24 &-&-&-&-\\
M2M-100  & 49.43 & 43.82 & 26.63 & 35.91 & 39.04 & 25.93 & 36.79\\
mBART  & 49.16 & 44.86 & 29.60 & 37.09 & 39.64 & - & -\\
MARGE$\dag$ & - & - & 28.40 &-&-&-&-\\
Baseline Model + BT &49.53&45.98&30.17 & 39.28 & 42.33 & 30.62 & 39.65\\
Multilingual Agreement & 48.99 & 47.34 &30.35&39.79&43.01&32.14 & 40.27\\
\hline
\multicolumn{8}{l}{\textit{Systems with Bilingual Parallel Document Data for Pre-training}}\\
\hline
DOCmT5$\dag$ & -&-&31.40$^*$&-&-&-&-\\
\hline
\hline
\textbf{BMA-SBT (Ours) + BT} & \textbf{51.10} & \textbf{47.59} & \textbf{30.80} & \textbf{40.20} & \textbf{43.17} & \textbf{32.23} & \textbf{40.85} \\
\hline
\hline
\multicolumn{8}{l}{\textit{Ablation Study}}\\
 - w/o $KL_1$ & 49.58 & 46.38 & 29.46 & 39.09 & 42.87 & 30.59 & 39.66\\
 - w/o $KL_2$ & 50.56 & 47.47 &30.26 & 40.02 & 43.15 & 31.89 & 40.56 \\
- w/o $KL_1$\&$KL_2$ &49.73&46.64&30.58&39.81 & 42.85 & 32.06 & 40.28 \\
\hline
\end{tabular}
\caption{\label{iwslt15_en}
Test results on TED Talks in the direction of (X $\rightarrow$ En). $\dag$: scores are taken from the official papers for these models. -: the scores are not reported or the language is not supported. *: the score is not directly comparable due to the use of document-level parallel corpora for pre-training. The Baseline Model refers to the model described in Section \ref{baselinemodel}, which is used for parameter initialization for BMA-SBT. BT refers to the conventional back-translation method described in Section \ref{bmabt-label}. $KL_1$ and $KL_2$ refers to the loss described in Equation \ref{eq5} and Equation \ref{eq6} respectively. We train our system BMA-SBT at the document level.
}
\end{table*}
where $\mathcal{L}_{KL_1}$ and $\mathcal{L}_{KL_2}$ represents the KL divergence loss in two directions:
\begin{equation}
\label{eq5}
    \mathbb{E}[KL(P_\theta(y\mid x)\mid\mid P_\theta(y\mid \tilde{z}))]
\end{equation}
for $\mathcal{L}_{KL_1}$, which means that we enforce the original source text $x$ to learn from the synthetic $\tilde{z}$. Note that $x$ and $\tilde{z}$ have the same meaning, but they are written in different languages. We also optimize in the other direction:
\begin{equation}
\label{eq6}
    \mathbb{E}[KL( P_\theta(y\mid \tilde{z})\mid\mid P_\theta(y\mid x))]
\end{equation}
for $\mathcal{L}_{KL_2}$.\footnote{Empirically, we have found that setting a balanced value with $\alpha=0.5$ brings a good performance.} In contrast to $KL_1$, this means that the synthetic text $\tilde{z}$ should learn from the original text $x$. Bidirectionality is necessary to enforce both languages to learn from each other. Here, $x$ and $y$ denote the original translation source and target respectively, and $\tilde{z}$ denotes the synthetic auxiliary text created by BMA-SBT via translation.
\paragraph{BMA-SBT}
Overall, we propose a novel BMA-SBT framework that optimizes the MNMT models with the following combinatory loss:
\begin{equation}
\label{eq4}
\mathcal{L}_{BMA-SBT}=\mathcal{L}_{main}+\mathcal{L}_{auxiliary}+\mathcal{L}_{BMA}
\end{equation}
Figure \ref{f2} depicts the overview of BMA-SBT. The final KL loss at the right edge of the figure refers to $\mathcal{L}_{BMA}$, $\mathcal{L}_{main}$ is calculated with the training instance at the top, and $\mathcal{L}_{auxiliary}$ is calculated with the training instance at the bottom.
\par 
BMA-SBT can be improved with multiple auxiliary languages for agreement in an ensemble manner. This requires more tuning and computational costs. We leave this to future work.
\section{Experiments}
\subsection{Implementation Details}
\label{baselinemodel}
\paragraph{Model Configuration}
The Transformer architecture we use is composed of 24 encoder layers and 12 interleaved decoder layers. Furthermore, the architecture has an embedding size of 1024, with a dropout rate of 0.1. The feed-forward network has a size of 4096, with 16 attention heads. For parameter initialization, we follow  \citet{2021arXiv210613736M} and \citet{yang-etal-2021-multilingual-machine} to pre-train a strong MT system with sentence-level bilingual data. For the rest of this paper, We call it the Baseline Model and use it as a strong baseline system.
\paragraph{Data Pre-processing}
For all of the experiments conducted in this paper, we use SentencePiece \citep{kudo-richardson-2018-sentencepiece} for tokenization. The SentencePiece model we use is the same as 
\citet{yang-etal-2021-multilingual-machine}. Also, we follow prior works to prefix the source input translation texts with a language tag that indicates the target language of the outputs.
\paragraph{Evaluations} We use the BLEU scores \citep{papineni-etal-2002-bleu} computed with the script from SacreBLEU for evaluation.\footnote{\url{https://github.com/mjpost/sacrebleu}}

\paragraph{Training Details}
We use the Adam optimizer \citep{Adam} and set it with the hyperparameter $\beta_1=0.9$ and $\beta_2=0.98$ for downstream fine-tuning. We set the learning rate as $1e\text{-}5$, with a warmup step of $4000$. We use the label smoothing cross-entropy for the standard translation loss and we set label smoothing with a ratio of $0.1$ for model training. All of the fine-tuning experiments reported in this paper are conducted on 8 NVIDIA V100 GPUs. We set the batch size as 512 tokens per GPU. Furthermore, to simulate a larger batch size, we update the models every 128 steps. For bilingual back-translation models, we use the downstream datasets for training on the same Transformer architecture.
\subsection{TED Talks}
\paragraph{Experimental Settings}
 We use the IWSLT15 Campaign for the evaluation of TED Talks, on the task of multilingual MT. Prior systems have reported scores on only 1 or 2 translation directions \citep{lee-etal-2022-docmt5, sun-etal-2022-rethinking}, and \citet{lee-etal-2022-docmt5} supports only the translation direction into English (X $\rightarrow$ En). We report a wider range of language directions on the benchmark. We split all documents into sub-documents with a maximum of 512 tokens for all train/dev/test sets during training and inference. We use the official parallel training data from IWSLT15 with no additional monolingual data and the official 2010 dev set and 2010-2013 test set for evaluation \citep{liu-etal-2020-multilingual-denoising,lee-etal-2022-docmt5}. We use the Baseline Model to generate all the BT data and the SBT data used for multilingual agreement in BMA-SBT. We fine-tune our model BMA-SBT at the document level. We report d-BLEU \citep{liu-etal-2020-multilingual-denoising} using SacreBLEU.\footnote{\url{https://github.com/mjpost/sacrebleu}} d-BLEU score is a BLEU score for documents.
 \begin{table*}[h!]
\centering
    \setlength\tabcolsep{6pt}
\setlength\aboverulesep{0pt}\setlength\belowrulesep{0pt}
\setcellgapes{2pt}\makegapedcells
\begin{tabular}{l|cccc|c}
\hline
\textbf{Model} & \,\textbf{Fr$\rightarrow$En}\, & \,\textbf{De$\rightarrow$En}\, & \,\textbf{Zh$\rightarrow$En}\, & \,\textbf{Cs $\rightarrow$En}\, & \,\textbf{Avg.}\,\\
\hline
\multicolumn{6}{l}{\textit{Sentence-level Systems}}
\\
\hline
M2M-100 \citep{10.5555/3546258.3546365} & 31.58 & 25.65 & 18.47 & 28.17 & 25.97\\
mBART \citep{liu-etal-2020-multilingual-denoising} & 29.93 & 29.31 & 18.33 & 30.15 & 26.93\\
\hline
\multicolumn{6}{l}{\textit{Document-level Systems}}\\
\hline
M2M-100 \citep{10.5555/3546258.3546365} & 32.67 & 25.78 & 17.85 & 29.06 & 26.34\\
mBART \citep{liu-etal-2020-multilingual-denoising} & 30.14 & 26.35 & 15.01 & 29.79 & 25.32\\
Baseline Model \citep{yang-etal-2021-multilingual-machine} + BT &36.38&34.24&25.58&36.97&33.29\\
\hline
\textbf{BMA-SBT (Ours) + BT} & \textbf{37.26} & \textbf{34.58} & \textbf{26.31} & \textbf{37.58}&\textbf{33.93}\\
\hline
\end{tabular}
\caption{\label{news_en}
Test results on the News benchmark in the direction of (X $\rightarrow$ En).
}
\end{table*}
\paragraph{Baseline Systems}
We report strong baselines evaluated at both sentence and document levels. Evaluating at the sentence level means that we split documents into sentences for training and inference. In contrast, evaluating at the document level means that we split all documents into sub-documents with a maximum of 512 tokens as described in the Experimental Settings. We compare to the following baselines: M2M-100 \citep{10.5555/3546258.3546365}, mBART \citep{liu-etal-2020-multilingual-denoising}, HAN$\dag$ \citep{yang-etal-2016-hierarchical}, MARGE$\dag$ \citep{10.5555/3495724.3497275}, and the Baseline Model that we use to initialize the weights for BMA-SBT. $\dag$: the scores are taken from existing papers. We also report performance with Multilingual Agreement \citep{magree} fine-tuned on Baseline Model with BT using synthetic parallel text. For a fair comparison, we do not directly compare to the SOTA model DOCmT5$\dag$ \citep{lee-etal-2022-docmt5}, as it uses a large amount of bilingual parallel document data for a document-level multilingual pre-training. The corpus used by DOCmT5 is not publicly available yet, and our methodology does not make use of such data. See Appendix \ref{npa} for the number of model parameters.
\paragraph{Results}
\begin{table*}
\small 
\centering
\setlength\tabcolsep{4pt}
\setlength\aboverulesep{0pt}\setlength\belowrulesep{0pt}
\setcellgapes{1pt}\makegapedcells
\begin{tabular}{l|ccccccccc}
\hline
\textbf{Model} & \textbf{Da$\rightarrow$En} & \textbf{De$\rightarrow$En} & \textbf{El$\rightarrow$En} & \textbf{Es$\rightarrow$En} & \textbf{Fr$\rightarrow$En} & \textbf{It$\rightarrow$En} & \textbf{Nl$\rightarrow$En} & \textbf{Pt$\rightarrow$En} & \textbf{Sv$\rightarrow$En}\\
\hline
\multicolumn{10}{l}{\textit{Sentence-level Systems}}
\\
\hline
M2M-100 & 50.40 & 47.38 &52.28&52.03&48.26 & 49.70 & 46.78 &49.84&52.34 \\
Baseline Model + BT  &48.94 & 47.25 & 53.46 & 50.57 & 47.68 & 49.49 & 45.95 & 50.65 & 52.77\\
\hline
\multicolumn{10}{l}{\textit{Document-level Systems}}\\
\hline
M2M-100 & 50.33 & 47.00 &52.24&52.14&48.13 & 49.71 & 46.65 &40.68&52.28 \\
Baseline Model + BT  & 49.85& 47.64 & 53.34 & 51.32 & 48.46 & 50.26 & 47.12 & 50.13 & 52.42\\
 \hline
\textbf{BMA-SBT (Ours) + BT} & \textbf{50.52} & \textbf{47.86} & \textbf{54.06} & \textbf{52.17} & \textbf{48.77} & \textbf{50.67} & \textbf{47.90} & \textbf{50.69} & \textbf{52.96}\\
\hline
\end{tabular}
\caption{\label{europarl_main}
Test results on the Europarl benchmark in the direction of (X $\rightarrow$ En).
}
\end{table*}
Table \ref{iwslt15_en} presents the evaluation results of TED Talks in the direction of (X $\rightarrow$ En). BMA-SBT clearly surpasses the baselines.  BMA-SBT surpasses the Baseline Model when both are fine-tuned at the document level by an average of 1.20 points in the score. BMA-SBT surpasses the Baseline Model fine-tuned at the sentence level by an average of 0.34 points in the score. Here, the Baseline Model fine-tuned at the document level is no better than that of the sentence level. We postulate that the underlying reason is that previous works have reported that directly optimizing the MNMT model at the document level can be challenging due to the long input problem \citep{gtrans3}. For a fair comparison, we add the conventional back-translation (BT) to both BMA-SBT and the Baseline Model. See Section \ref{bmabt-label} for more explanation on the difference between BT and the SBT methods used to achieve multilingual agreement.  
\par
In addition to the fact that BMA-SBT clearly improves the Baseline Model, which is a strong pre-trained MT system, BMA-SBT also beats other baselines such as HAN, M2M-100, mT5, and mBART, both fine-tuned at the sentence level and at the document level. Indeed, the Baseline Model itself is already quite competitive with these models, and being able to improve such a model is a piece of clear evidence for the effectiveness of BMA-SBT. The final results we obtain are close to the SOTA system DOCmT5, which uses a large amount of bilingual document translation pairs for multilingual pre-training.
\paragraph{Ablation Study} The ablation study in Table \ref{iwslt15_en} supports three points of view: (i) the bidirectionality of the multilingual agreement is necessary, (ii) the synthetic additional parallel data created by the BT used for MA is useful, and (iii) BMA-SBT brings additional improvements to the BT.
\par
\begin{table*}[t!]
\scriptsize
\centering
\setlength\aboverulesep{0pt}\setlength\belowrulesep{0pt}
\setcellgapes{1.5pt}\makegapedcells
    \setlength\tabcolsep{7pt}
    \setlength\extrarowheight{0pt}
\begin{tabular}{l|p{10.5cm}}
\hline
\textbf{Source} & \begin{CJK*}{UTF8}{gbsn}……，当光在\underline{西红柿}上走过时，它一直在闪耀。它并没有变暗。为什么？因为\underline{西红柿}熟了，并且光在西红柿内部反射，
……
\end{CJK*}\\
\hline
\textbf{Reference} & ..., as the light washes over \hlc[cyan]{the tomato}, It continues to glow. It doesn't become dark. Why is that? Because \hlc[cyan]{the tomato is actually ripe}, and  \hlc[cyan]{the light} is bouncing around inside the tomato, ...\\
\hline
\textbf{Google Translate} & ..., as the light passed over \hlc[lime]{the tomatoes}, It kept shining. It didn't get darker. Why? Because \hlc[lime]{the tomatoes are ripe}, and \hlc[lime]{light} is reflected inside the tomatoes, ... \\
\hline
\textbf{Microsoft Translator} & ..., as the light walks over \hlc[lime]{the tomatoes}, It keeps shining. It didn't darken. Why? Because \hlc[lime]{the tomatoes are ripe}, and \hlc[lime]{light} is reflected inside the tomatoes, ... \\
\hline
\textbf{DeepL Translate} & ..., as the light traveled over \hlc[lime]{the tomatoes}, it kept shining. It doesn't dim. Why? Because \hlc[lime]{the tomatoes are ripe} and \hlc[cyan]{the light} is reflecting inside the tomatoes, ...\\
\hline
\textbf{Baseline Model (Sentence-level)} & ..., as the light goes over \hlc[cyan]{the tomato}, It's always glowing. It's not darkening. Why? Because \hlc[cyan]{the tomato is ripe}, and  \hlc[lime]{light} is reflected inside the tomato, ... \\
\hline
\textbf{Baseline Model (Document-level)} & ..., as the light passes over \hlc[cyan]{the tomato}, It keeps flashing. It doesn't get darker. Why? Because \hlc[lime]{the tomatoes are ripe} , and  \hlc[cyan]{the light} is is reflected inside the tomato, ... \\

\hline
\hline
\textbf{BMA-SBT} & ..., as the light passes over \hlc[cyan]{the tomato}, It's always shining. It's not darkening. Why? Because \hlc[cyan]{the tomato is ripe}, and \hlc[cyan]{the light} is reflected inside the tomato, ... \\
\hline
\end{tabular}
\caption{\label{case}
A Chinese-to-English case study from TED Talks demonstrates that BMA-SBT captures better noun-related issues. We highlight the correct translation in cyan (darker one when printed in B\&W), and the mistakes in lime (lighter one when printed in B\&W). Google Translate: \url{https://translate.google.com/}, Microsoft Translator: \url{https://www.bing.com/translator}, DeepL Translate: \url{https://www.deepl.com/translator}.
}
\end{table*}
Firstly, the row of (-w/o $KL_1$) and the row of (-w/o $KL_2$) represent the ablations when the KL loss in the directions described in Equation \ref{eq5} and Equation \ref{eq6} are ablated respectively. Here, we can see that both lead to a degradation in the results. Clearly, using $KL_2$ solely without $KL_1$ seems to degrade the performance. This is not surprising, as $KL_1$ pushes the output distributions of authentic data to be close to that of auxiliary text, which helps the model to use more linguistic clues in the auxiliary text. Also, using $KL_2$ solely pushes the outputs of synthetic auxiliary data to be close to that of the authentic data unidirectionally, which can be less helpful to the original authentic data. Removing $KL_2$ and using $KL_1$ solely also degrades the results, which aligns with our original motivation depicted for the bidirectionality as in Figure \ref{f1}.
\par
Secondly, the row of (- w/o $KL_1$\&$KL_2$) brings improvements compared to Baseline Model + BT, which means that the auxiliary parallel data itself created by switched back-translation is useful.
\par
Finally, BMA-SBT + BT brings clear improvements to the Baseline Model + BT. Since both models have used the conventional BT (See Section \ref{baselinemodel} for more details), the comparison is fair, which means that the BMA-SBT framework is effective and brings additive improvement to BT.
\subsection{News}
\paragraph{Experimental Settings} For evaluation on the News benchmark, we use News Commentary v11 as the training set, following \citet{sun-etal-2022-rethinking}. We employ newstest2015 as the dev set, and newstest2016/newstest2019 as the test set respectively for Cs and De. We use newstest2013 as the dev set and newstest2015 as the test set for Fr. We use newstest2019 as the dev set and newstest2020 as the test set for Zh. The remaining settings follow the same as the evaluation on TED Talks.
\paragraph{Baseline Systems}
As the weights for DOCmT5 are not available at the time of writing, we compare our system to various strong baselines such as M2M-100, mBART and the Baseline Model. We run the fine-tuning process on the official checkpoints to obtain the scores. For a fair comparison, we apply BT to the Baseline Model.
\paragraph{Results} Table \ref{news_en} compares BMA-SBT to strong baselines, and we see that the improvements with BMA-SBT are clear, and the final results surpass all the strong baselines. This validates BMA-SBT's effectiveness as a novel framework.
\subsection{Europarl}
\paragraph{Experimental Settings}
For the Europarl dataset \citep{koehn-2005-europarl}, we use Europarl-v7 \citet{sun-etal-2022-rethinking}. W experiment with (X $\rightarrow$ En) where we test nine languages: Da, De, El, Es, Fr, It, Nl, Pt, and Sv. Like previous works \citep{bao-etal-2021-g, sun-etal-2022-rethinking}, the dataset is randomly partitioned into train/dev/test divisions, and we split by English document IDs to avoid information leakage to better support the multilingual setting.
\paragraph{Baseline Systems} As the weights for DOCmT5 are not available at the time of writing, we compare our system to various strong baselines such as M2M-100 and the Baseline Model. We run the fine-tuning process on the official checkpoints to obtain the scores. For a fair comparison, we apply BT to the Baseline Model.
\paragraph{Results} Table \ref{europarl_main} compares BMA-SBT to strong baselines, and we see that the improvements with BMA-SBT are obvious, and the final results surpass all the strong baselines.
\subsection{Case Study} Table \ref{case} depicts a Zh$\rightarrow$En case study on TED Talks. In addition to the Baseline Models, we also compare BMA-SBT to various commercial systems such as Google Translate.  In this case, we see that the Chinese text does not differentiate plural from single. Among all cases, it is clear that BMA-SBT works the best and can effectively resolve such ambiguity. We also observe that BMA-SBT perfectly capture the context and attaches the definite article `the' to `light'. This aligns with our original intention depicted in Figure \ref{f1} to help the models to improve cross-lingual learning via BMA-SBT.
\subsection{Coherence and Consistency Evaluation} 
\begin{figure}[t!]
\begin{center}
\centerline{
\includegraphics[width=9cm]{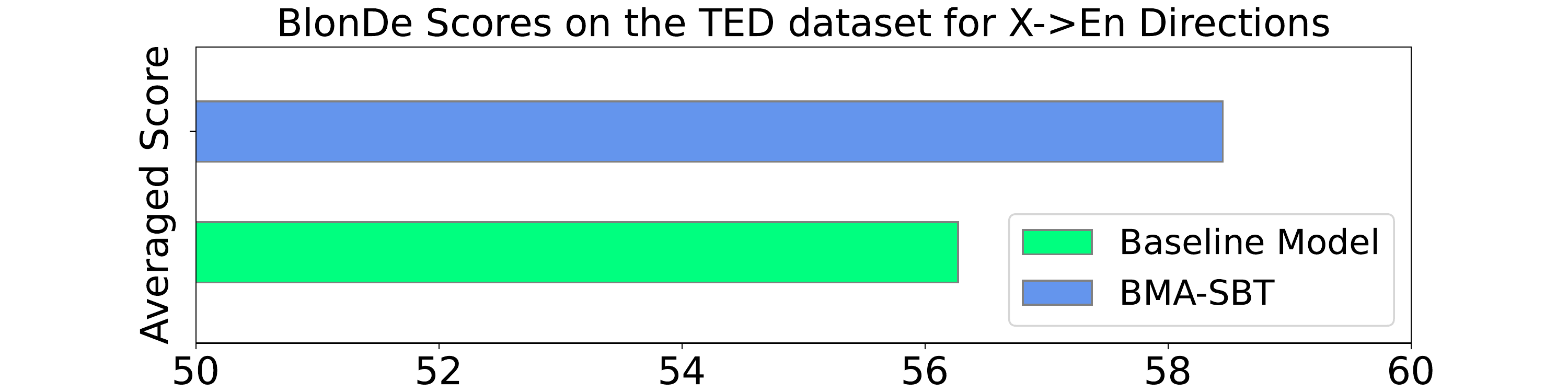}}
\caption{Averaged BlonDe scores from six directions in (X $\rightarrow$ En) on the dataset of TED Talks evaluated with BMA-SBT and the Baseline Model (Document-level).
}
\label{blonde}
\end{center}
\vspace{-5mm}
\end{figure}

Figure \ref{blonde} depicts the evaluations in the averaged scores from six translation directions on TED Talks with BlonDe scores \citep{blonde}. BlonDe is an evaluation metric designed for MT which considers document-level coherence and consistency issues that require the model to resolve cross-lingual ambiguities. We see that BMA-SBT brings effective improvements to the metric.
\section{Related Work}
\subsection{Multilingual Neural Machine Translation}
Conventional bilingual machine translation models deal with two languages: one as the input, and one as the output. In comparison, multilingual neural machine translation (MNMT) has achieved great success in handling multiple languages with a single model. Recently, there have been many pre-training works on MNMT through multilingual pre-training models that leverage unsupervised pre-training objectives on monolingual corpora in many different languages \citep{conneau-etal-2020-unsupervised,liu-etal-2020-multilingual-denoising, xue-etal-2021-mt5}. Following the calls that the unsupervised scenario is not strictly realistic for cross-lingual learning \citep{artetxe-etal-2020-call}, subsequent works use parallel corpora with translation pairs for multilingual pre-training \citep{ reid-artetxe-2022-paradise, lee-etal-2022-docmt5}.
\par
While pre-training has shown great success for MNMT \citep{nllb2022}, it is unclear whether the previous methods for from-scratching training on MNMT are still useful on pre-trained models. Multilingual agreement \citep{magree} is perhaps the closest work to ours among those methods for from-scratch training. However, conventional MA requires authentic parallel data among many language pairs, which does not always guarantee to exist. In comparison, we focus on a more recent fine-tuning setting on popular pre-training models as well as a realistic setting with no presumption on the existence of the additional parallel data.
\subsection{Agreement-based Learning}
Agreement-based learning has been proven as a useful paradigm in the language community \citep{agreement1, agreement2, agreement5,NEURIPS2021_5a66b9205}. The core idea is to minimize the difference in the representations between the inputs with the same meaning. Some multilingual pre-training methods such as \citet{chi-etal-2021-infoxlm} are relevant to agreement-based learning in the way that they shrink the distance of cross-lingual representations between parallel data.
\citet{agreement3} proposed to enforce an agreement on the output with left-to-right and right-to-left inputs on recurrent neural networks for machine translation. \citet{agreement4} proposed to use phrase-level agreement for machine translation.
\par
Still, \citet{magree} is the closest work to ours, which encourages agreement between parallel data in different languages to have the same translation outputs. Our work creates synthetic data and employs bidirectional KL loss to enforce the multilingual agreement bidirectionally.
\section{Conclusions} Despite the fact that multilingual agreement (MA) has shown its effectiveness in from-scratch training for MNMT, the conventional MA has at least two shortages that limit its usages: (i) needs authentic extra parallel data, which can be often unrealistic and (ii) has an ambiguous direction for agreement-based learning. We propose BMA-SBT as a novel and universal fine-tuning framework for pre-trained MT models that (i) exempts the need for authentic parallel data by creating synthetic parallel text written in a different source language and (ii) specifies the direction of agreement-based learning with bidirectional KL divergence loss. Experimental results on three multilingual machine translation datasets illustrate that BMA-SBT can obviously improve the strong pre-trained baseline system. An in-depth investigation indicates that BMA-SBT brings additive improvements to the conventional BT methods for neural machine translation.
\section*{Limitations}
The proposed method requires generating synthetic auxiliary parallel data using translation models, which requires extra computational costs. The proposed method requires generating synthetic auxiliary parallel data using translation models, which requires extra computational costs. 
\paragraph{Large Language Models} Large Language Models (LLMs) such as ChatGPT have shown good translation abilities \citep{2023arXiv230506575L}, while they still lag behind supervised systems \citep{2023arXiv230108745J, 2023arXiv230404675Z}. We do not directly compare them, as they are much larger in their number of parameters than the systems described in this work.  
\section*{Ethics Statement}
We honour and support the EMNLP Code of Ethics. The datasets used in this work are well-known and widely used, and the dataset pre-processing does not make use of any external textual resource. In our view, there is no known ethical issue. End-to-end pre-trained generators are also used, which are subjected to generating offensive context. But the above-mentioned issues are widely known to commonly exist for these models. Any content generated does not reflect the view of the authors.
\bibliography{anthology,custom}

\begin{thebibliography}{35}
\expandafter\ifx\csname natexlab\endcsname\relax\def\natexlab#1{#1}\fi

\bibitem[{Artetxe et~al.(2020)Artetxe, Ruder, Yogatama, Labaka, and
  Agirre}]{artetxe-etal-2020-call}
Mikel Artetxe, Sebastian Ruder, Dani Yogatama, Gorka Labaka, and Eneko Agirre.
  2020.
\newblock \href {https://doi.org/10.18653/v1/2020.acl-main.658} {A call for
  more rigor in unsupervised cross-lingual learning}.
\newblock In \emph{Proceedings of the 58th Annual Meeting of the Association
  for Computational Linguistics}, pages 7375--7388, Online. Association for
  Computational Linguistics.

\bibitem[{Bao et~al.(2021)Bao, Zhang, Teng, Chen, and Luo}]{bao-etal-2021-g}
Guangsheng Bao, Yue Zhang, Zhiyang Teng, Boxing Chen, and Weihua Luo. 2021.
\newblock \href {https://doi.org/10.18653/v1/2021.acl-long.267}
  {{G}-transformer for document-level machine translation}.
\newblock In \emph{Proceedings of the 59th Annual Meeting of the Association
  for Computational Linguistics and the 11th International Joint Conference on
  Natural Language Processing (Volume 1: Long Papers)}, pages 3442--3455,
  Online. Association for Computational Linguistics.

\bibitem[{Cettolo et~al.(2015)Cettolo, Niehues, St{\"u}ker, Bentivogli,
  Cattoni, and Federico}]{cettolo-etal-2015-iwslt}
Mauro Cettolo, Jan Niehues, Sebastian St{\"u}ker, Luisa Bentivogli, Roldano
  Cattoni, and Marcello Federico. 2015.
\newblock \href {https://aclanthology.org/2015.iwslt-evaluation.1} {The {IWSLT}
  2015 evaluation campaign}.
\newblock In \emph{Proceedings of the 12th International Workshop on Spoken
  Language Translation: Evaluation Campaign}, pages 2--14, Da Nang, Vietnam.

\bibitem[{Cheng et~al.(2016)Cheng, Shen, He, He, Wu, Sun, and Liu}]{agreement5}
Yong Cheng, Shiqi Shen, Zhongjun He, Wei He, Hua Wu, Maosong Sun, and Yang Liu.
  2016.
\newblock Agreement-based joint training for bidirectional attention-based
  neural machine translation.
\newblock In \emph{Proceedings of the Twenty-Fifth International Joint
  Conference on Artificial Intelligence}, IJCAI'16, page 2761–2767. AAAI
  Press.

\bibitem[{Chi et~al.(2021)Chi, Dong, Wei, Yang, Singhal, Wang, Song, Mao,
  Huang, and Zhou}]{chi-etal-2021-infoxlm}
Zewen Chi, Li~Dong, Furu Wei, Nan Yang, Saksham Singhal, Wenhui Wang, Xia Song,
  Xian-Ling Mao, Heyan Huang, and Ming Zhou. 2021.
\newblock \href {https://doi.org/10.18653/v1/2021.naacl-main.280} {{I}nfo{XLM}:
  An information-theoretic framework for cross-lingual language model
  pre-training}.
\newblock In \emph{Proceedings of the 2021 Conference of the North American
  Chapter of the Association for Computational Linguistics: Human Language
  Technologies}, pages 3576--3588, Online. Association for Computational
  Linguistics.

\bibitem[{Conneau et~al.(2020)Conneau, Khandelwal, Goyal, Chaudhary, Wenzek,
  Guzm{\'a}n, Grave, Ott, Zettlemoyer, and
  Stoyanov}]{conneau-etal-2020-unsupervised}
Alexis Conneau, Kartikay Khandelwal, Naman Goyal, Vishrav Chaudhary, Guillaume
  Wenzek, Francisco Guzm{\'a}n, Edouard Grave, Myle Ott, Luke Zettlemoyer, and
  Veselin Stoyanov. 2020.
\newblock \href {https://doi.org/10.18653/v1/2020.acl-main.747} {Unsupervised
  cross-lingual representation learning at scale}.
\newblock In \emph{Proceedings of the 58th Annual Meeting of the Association
  for Computational Linguistics}, pages 8440--8451, Online. Association for
  Computational Linguistics.

\bibitem[{Fan et~al.(2022)Fan, Bhosale, Schwenk, Ma, El-Kishky, Goyal, Baines,
  Celebi, Wenzek, Chaudhary, Goyal, Birch, Liptchinsky, Edunov, Grave, Auli,
  and Joulin}]{10.5555/3546258.3546365}
Angela Fan, Shruti Bhosale, Holger Schwenk, Zhiyi Ma, Ahmed El-Kishky,
  Siddharth Goyal, Mandeep Baines, Onur Celebi, Guillaume Wenzek, Vishrav
  Chaudhary, Naman Goyal, Tom Birch, Vitaliy Liptchinsky, Sergey Edunov,
  Edouard Grave, Michael Auli, and Armand Joulin. 2022.
\newblock Beyond english-centric multilingual machine translation.
\newblock \emph{J. Mach. Learn. Res.}, 22(1).

\bibitem[{Jiang et~al.(2022)Jiang, Liu, Ma, Zhang, Yang, Huang, Sennrich,
  Cotterell, Sachan, and Zhou}]{blonde}
Yuchen Jiang, Tianyu Liu, Shuming Ma, Dongdong Zhang, Jian Yang, Haoyang Huang,
  Rico Sennrich, Ryan Cotterell, Mrinmaya Sachan, and Ming Zhou. 2022.
\newblock \href {https://doi.org/10.18653/v1/2022.naacl-main.111} {{BlonDe}: An
  automatic evaluation metric for document-level machine translation}.
\newblock In \emph{Proceedings of the 2022 Conference of the North American
  Chapter of the Association for Computational Linguistics: Human Language
  Technologies}, pages 1550--1565, Seattle, United States. Association for
  Computational Linguistics.

\bibitem[{{Jiao} et~al.(2023){Jiao}, {Wang}, {Huang}, {Wang}, and
  {Tu}}]{2023arXiv230108745J}
Wenxiang {Jiao}, Wenxuan {Wang}, Jen-tse {Huang}, Xing {Wang}, and Zhaopeng
  {Tu}. 2023.
\newblock \href {https://doi.org/10.48550/arXiv.2301.08745} {{Is ChatGPT A Good
  Translator? Yes With GPT-4 As The Engine}}.
\newblock \emph{arXiv e-prints}, page arXiv:2301.08745.

\bibitem[{Kingma and Ba(2014)}]{Adam}
Diederik Kingma and Jimmy Ba. 2014.
\newblock Adam: A method for stochastic optimization.
\newblock \emph{International Conference on Learning Representations}.

\bibitem[{Koehn(2005)}]{koehn-2005-europarl}
Philipp Koehn. 2005.
\newblock \href {https://aclanthology.org/2005.mtsummit-papers.11} {{E}uroparl:
  A parallel corpus for statistical machine translation}.
\newblock In \emph{Proceedings of Machine Translation Summit X: Papers}, pages
  79--86, Phuket, Thailand.

\bibitem[{Koehn and Knowles(2017)}]{gtrans3}
Philipp Koehn and Rebecca Knowles. 2017.
\newblock \href {https://doi.org/10.18653/v1/W17-3204} {Six challenges for
  neural machine translation}.
\newblock In \emph{Proceedings of the First Workshop on Neural Machine
  Translation}, pages 28--39, Vancouver. Association for Computational
  Linguistics.

\bibitem[{Kudo and Richardson(2018)}]{kudo-richardson-2018-sentencepiece}
Taku Kudo and John Richardson. 2018.
\newblock \href {https://doi.org/10.18653/v1/D18-2012} {{S}entence{P}iece: A
  simple and language independent subword tokenizer and detokenizer for neural
  text processing}.
\newblock In \emph{Proceedings of the 2018 Conference on Empirical Methods in
  Natural Language Processing: System Demonstrations}, pages 66--71, Brussels,
  Belgium. Association for Computational Linguistics.

\bibitem[{Lee et~al.(2022)Lee, Siddhant, Ratnakar, and
  Johnson}]{lee-etal-2022-docmt5}
Chia-Hsuan Lee, Aditya Siddhant, Viresh Ratnakar, and Melvin Johnson. 2022.
\newblock \href {https://doi.org/10.18653/v1/2022.findings-naacl.32}
  {{DOC}m{T}5: Document-level pretraining of multilingual language models}.
\newblock In \emph{Findings of the Association for Computational Linguistics:
  NAACL 2022}, pages 425--437, Seattle, United States. Association for
  Computational Linguistics.

\bibitem[{Lewis et~al.(2020)Lewis, Ghazvininejad, Ghosh, Aghajanyan, Wang, and
  Zettlemoyer}]{10.5555/3495724.3497275}
Mike Lewis, Marjan Ghazvininejad, Gargi Ghosh, Armen Aghajanyan, Sida Wang, and
  Luke Zettlemoyer. 2020.
\newblock Pre-training via paraphrasing.
\newblock In \emph{Proceedings of the 34th International Conference on Neural
  Information Processing Systems}, NIPS'20, Red Hook, NY, USA. Curran
  Associates Inc.

\bibitem[{Liang et~al.(2007)Liang, Klein, and Jordan}]{agreement2}
Percy Liang, Dan Klein, and Michael~I. Jordan. 2007.
\newblock \href
  {https://proceedings.neurips.cc/paper/2007/hash/dbe272bab69f8e13f14b405e038deb64-Abstract.html}
  {Agreement-based learning}.
\newblock In \emph{Advances in Neural Information Processing Systems 20,
  Proceedings of the Twenty-First Annual Conference on Neural Information
  Processing Systems, Vancouver, British Columbia, Canada, December 3-6, 2007},
  pages 913--920. Curran Associates, Inc.

\bibitem[{Liang et~al.(2006)Liang, Taskar, and Klein}]{agreement1}
Percy Liang, Ben Taskar, and Dan Klein. 2006.
\newblock \href {https://aclanthology.org/N06-1014} {Alignment by agreement}.
\newblock In \emph{Proceedings of the Human Language Technology Conference of
  the {NAACL}, Main Conference}, pages 104--111, New York City, USA.
  Association for Computational Linguistics.

\bibitem[{Liu et~al.(2020)Liu, Gu, Goyal, Li, Edunov, Ghazvininejad, Lewis, and
  Zettlemoyer}]{liu-etal-2020-multilingual-denoising}
Yinhan Liu, Jiatao Gu, Naman Goyal, Xian Li, Sergey Edunov, Marjan
  Ghazvininejad, Mike Lewis, and Luke Zettlemoyer. 2020.
\newblock \href {https://doi.org/10.1162/tacl_a_00343} {Multilingual denoising
  pre-training for neural machine translation}.
\newblock \emph{Transactions of the Association for Computational Linguistics},
  8:726--742.

\bibitem[{{Lu} et~al.(2022){Lu}, {Huang}, {Ma}, {Zhang}, {Lam}, and
  {Wei}}]{2022arXiv221207752L}
Hongyuan {Lu}, Haoyang {Huang}, Shuming {Ma}, Dongdong {Zhang}, Wai {Lam}, and
  Furu {Wei}. 2022.
\newblock \href {https://doi.org/10.48550/arXiv.2212.07752} {{TRIP: Triangular
  Document-level Pre-training for Multilingual Language Models}}.
\newblock \emph{arXiv e-prints}, page arXiv:2212.07752.

\bibitem[{{Lu} et~al.(2023){Lu}, {Huang}, {Zhang}, {Yang}, {Lam}, and
  {Wei}}]{2023arXiv230506575L}
Hongyuan {Lu}, Haoyang {Huang}, Dongdong {Zhang}, Haoran {Yang}, Wai {Lam}, and
  Furu {Wei}. 2023.
\newblock \href {http://arxiv.org/abs/2305.06575} {{Chain-of-Dictionary
  Prompting Elicits Translation in Large Language Models}}.
\newblock \emph{arXiv e-prints}, page arXiv:2305.06575.

\bibitem[{{Ma} et~al.(2021){Ma}, {Dong}, {Huang}, {Zhang}, {Muzio}, {Singhal},
  {Hassan Awadalla}, {Song}, and {Wei}}]{2021arXiv210613736M}
Shuming {Ma}, Li~{Dong}, Shaohan {Huang}, Dongdong {Zhang}, Alexandre {Muzio},
  Saksham {Singhal}, Hany {Hassan Awadalla}, Xia {Song}, and Furu {Wei}. 2021.
\newblock \href {http://arxiv.org/abs/2106.13736} {{DeltaLM: Encoder-Decoder
  Pre-training for Language Generation and Translation by Augmenting Pretrained
  Multilingual Encoders}}.
\newblock \emph{arXiv e-prints}, page arXiv:2106.13736.

\bibitem[{NLLB-Team(2022)}]{nllb2022}
NLLB-Team. 2022.
\newblock No language left behind: Scaling human-centered machine translation.

\bibitem[{Papineni et~al.(2002)Papineni, Roukos, Ward, and
  Zhu}]{papineni-etal-2002-bleu}
Kishore Papineni, Salim Roukos, Todd Ward, and Wei-Jing Zhu. 2002.
\newblock \href {https://doi.org/10.3115/1073083.1073135} {{B}leu: a method for
  automatic evaluation of machine translation}.
\newblock In \emph{Proceedings of the 40th Annual Meeting of the Association
  for Computational Linguistics}, pages 311--318, Philadelphia, Pennsylvania,
  USA. Association for Computational Linguistics.

\bibitem[{Reid and Artetxe(2022)}]{reid-artetxe-2022-paradise}
Machel Reid and Mikel Artetxe. 2022.
\newblock \href {https://doi.org/10.18653/v1/2022.naacl-main.58} {{PARADISE}:
  Exploiting parallel data for multilingual sequence-to-sequence pretraining}.
\newblock In \emph{Proceedings of the 2022 Conference of the North American
  Chapter of the Association for Computational Linguistics: Human Language
  Technologies}, pages 800--810, Seattle, United States. Association for
  Computational Linguistics.

\bibitem[{Sennrich et~al.(2016)Sennrich, Haddow, and
  Birch}]{sennrich-etal-2016-improving}
Rico Sennrich, Barry Haddow, and Alexandra Birch. 2016.
\newblock \href {https://doi.org/10.18653/v1/P16-1009} {Improving neural
  machine translation models with monolingual data}.
\newblock In \emph{Proceedings of the 54th Annual Meeting of the Association
  for Computational Linguistics (Volume 1: Long Papers)}, pages 86--96, Berlin,
  Germany. Association for Computational Linguistics.

\bibitem[{Sun et~al.(2022)Sun, Wang, Zhou, Zhao, Huang, Chen, and
  Li}]{sun-etal-2022-rethinking}
Zewei Sun, Mingxuan Wang, Hao Zhou, Chengqi Zhao, Shujian Huang, Jiajun Chen,
  and Lei Li. 2022.
\newblock \href {https://doi.org/10.18653/v1/2022.findings-acl.279} {Rethinking
  document-level neural machine translation}.
\newblock In \emph{Findings of the Association for Computational Linguistics:
  ACL 2022}, pages 3537--3548, Dublin, Ireland. Association for Computational
  Linguistics.

\bibitem[{Sutskever et~al.(2014)Sutskever, Vinyals, and Le}]{S2S}
Ilya Sutskever, Oriol Vinyals, and Quoc~V. Le. 2014.
\newblock Sequence to sequence learning with neural networks.
\newblock In \emph{Proceedings of the 27th International Conference on Neural
  Information Processing Systems - Volume 2}, NIPS’14, page 3104–3112,
  Cambridge, MA, USA. MIT Press.

\bibitem[{Xue et~al.(2021)Xue, Constant, Roberts, Kale, Al-Rfou, Siddhant,
  Barua, and Raffel}]{xue-etal-2021-mt5}
Linting Xue, Noah Constant, Adam Roberts, Mihir Kale, Rami Al-Rfou, Aditya
  Siddhant, Aditya Barua, and Colin Raffel. 2021.
\newblock \href {https://doi.org/10.18653/v1/2021.naacl-main.41} {m{T}5: A
  massively multilingual pre-trained text-to-text transformer}.
\newblock In \emph{Proceedings of the 2021 Conference of the North American
  Chapter of the Association for Computational Linguistics: Human Language
  Technologies}, pages 483--498, Online. Association for Computational
  Linguistics.

\bibitem[{Yang et~al.(2021{\natexlab{a}})Yang, Ma, Huang, Zhang, Dong, Huang,
  Muzio, Singhal, Hassan, Song, and Wei}]{tulg1}
Jian Yang, Shuming Ma, Haoyang Huang, Dongdong Zhang, Li~Dong, Shaohan Huang,
  Alexandre Muzio, Saksham Singhal, Hany Hassan, Xia Song, and Furu Wei.
  2021{\natexlab{a}}.
\newblock \href {https://aclanthology.org/2021.wmt-1.54} {Multilingual machine
  translation systems from {M}icrosoft for {WMT}21 shared task}.
\newblock In \emph{Proceedings of the Sixth Conference on Machine Translation},
  pages 446--455, Online. Association for Computational Linguistics.

\bibitem[{Yang et~al.(2021{\natexlab{b}})Yang, Ma, Huang, Zhang, Dong, Huang,
  Muzio, Singhal, Hassan, Song, and Wei}]{yang-etal-2021-multilingual-machine}
Jian Yang, Shuming Ma, Haoyang Huang, Dongdong Zhang, Li~Dong, Shaohan Huang,
  Alexandre Muzio, Saksham Singhal, Hany Hassan, Xia Song, and Furu Wei.
  2021{\natexlab{b}}.
\newblock \href {https://aclanthology.org/2021.wmt-1.54} {Multilingual machine
  translation systems from {M}icrosoft for {WMT}21 shared task}.
\newblock In \emph{Proceedings of the Sixth Conference on Machine Translation},
  pages 446--455, Online. Association for Computational Linguistics.

\bibitem[{Yang et~al.(2021{\natexlab{c}})Yang, Yin, Ma, Huang, Zhang, Li, and
  Wei}]{magree}
Jian Yang, Yuwei Yin, Shuming Ma, Haoyang Huang, Dongdong Zhang, Zhoujun Li,
  and Furu Wei. 2021{\natexlab{c}}.
\newblock \href {https://doi.org/10.18653/v1/2021.acl-short.31} {Multilingual
  agreement for multilingual neural machine translation}.
\newblock In \emph{Proceedings of the 59th Annual Meeting of the Association
  for Computational Linguistics and the 11th International Joint Conference on
  Natural Language Processing (Volume 2: Short Papers)}, pages 233--239,
  Online. Association for Computational Linguistics.

\bibitem[{Yang et~al.(2020)Yang, Wang, Zhang, and Zhao}]{agreement4}
Mingming Yang, Xing Wang, Min Zhang, and Tiejun Zhao. 2020.
\newblock \href {https://doi.org/10.1007/978-3-030-60450-9_33} {Incorporating
  phrase-level agreement into neural machine translation}.
\newblock In \emph{Natural Language Processing and Chinese Computing: 9th CCF
  International Conference, NLPCC 2020, Zhengzhou, China, October 14–18,
  2020, Proceedings, Part I}, page 416–428, Berlin, Heidelberg.
  Springer-Verlag.

\bibitem[{Yang et~al.(2016)Yang, Yang, Dyer, He, Smola, and
  Hovy}]{yang-etal-2016-hierarchical}
Zichao Yang, Diyi Yang, Chris Dyer, Xiaodong He, Alex Smola, and Eduard Hovy.
  2016.
\newblock \href {https://doi.org/10.18653/v1/N16-1174} {Hierarchical attention
  networks for document classification}.
\newblock In \emph{Proceedings of the 2016 Conference of the North {A}merican
  Chapter of the Association for Computational Linguistics: Human Language
  Technologies}, pages 1480--1489, San Diego, California. Association for
  Computational Linguistics.

\bibitem[{Zhang et~al.(2019)Zhang, Wu, Liu, Li, Zhou, and Xu}]{agreement3}
Zhirui Zhang, Shuangzhi Wu, Shujie Liu, Mu~Li, Ming Zhou, and Tong Xu. 2019.
\newblock \href {https://doi.org/10.1609/aaai.v33i01.3301443} {Regularizing
  neural machine translation by target-bidirectional agreement}.
\newblock In \emph{Proceedings of the Thirty-Third AAAI Conference on
  Artificial Intelligence and Thirty-First Innovative Applications of
  Artificial Intelligence Conference and Ninth AAAI Symposium on Educational
  Advances in Artificial Intelligence}, AAAI'19/IAAI'19/EAAI'19. AAAI Press.

\bibitem[{{Zhu} et~al.(2023){Zhu}, {Liu}, {Dong}, {Xu}, {Huang}, {Kong},
  {Chen}, and {Li}}]{2023arXiv230404675Z}
Wenhao {Zhu}, Hongyi {Liu}, Qingxiu {Dong}, Jingjing {Xu}, Shujian {Huang},
  Lingpeng {Kong}, Jiajun {Chen}, and Lei {Li}. 2023.
\newblock \href {https://doi.org/10.48550/arXiv.2304.04675} {{Multilingual
  Machine Translation with Large Language Models: Empirical Results and
  Analysis}}.
\newblock \emph{arXiv e-prints}, page arXiv:2304.04675.

\end{thebibliography}
\bibliographystyle{acl_natbib}
\newpage
\newpage
\appendix
\section{Number of Model Parameters}
\label{npa}
\begin{table}[htb]
\centering
\setlength\tabcolsep{7pt}
\setlength\aboverulesep{0pt}\setlength\belowrulesep{0pt}
\setcellgapes{3pt}\makegapedcells
\begin{tabular}{l|c}
\hline
\textbf{Model} & \textbf{Number of Parameters}\\
\hline
M2M-100 & 418M\\
mBART & 611M\\
MARGE & 963M\\
mT5 & 1.23B$^*$\\
DOCmT5 & 1.23B$^*$\\
Baseline Model & 862M\\
\hline
\textbf{BMA-SBT (Ours)} & 862M\\

\hline
\end{tabular}
\caption{\label{np}
Comparison in the number of parameters for the pre-trained models used in our experiments. $*$: these models all use the model architecture of mT5-Large, and we report the number of model parameters taken from the original paper of mT5 reported by \citet{xue-etal-2021-mt5}.
}
\end{table}
Table \ref{np} presents the number of model parameters for the pre-trained models used in our experiments.
\end{document}